\documentclass{llncs}

\usepackage{ifpdf}
\usepackage{times}

\usepackage{cite}
\usepackage{graphicx}

\usepackage{amsmath}
\usepackage{amsfonts}
\usepackage{mathrsfs}
\usepackage{dsfont}
\usepackage{amsbsy}
\usepackage{stmaryrd}

\usepackage{algorithm}
\usepackage{algorithmic}
\usepackage{xpatch}
\usepackage{caption}
\usepackage{graphicx}
\usepackage[tight,footnotesize]{subfigure}
\usepackage{url}
\usepackage{tikz}
\usetikzlibrary{arrows,positioning,automata,circuits.logic.US,circuits.logic.IEC}

\xpatchcmd{\algorithmic}{\setcounter}{\algorithmicfont\setcounter}{}{}
\providecommand{\algorithmicfont}{}

\newcommand{\bb}[1]{\mathbb{#1}}
\addtocounter{page}{0}

\newcommand{\sspace}[1]{D_{#1}}

\renewcommand{\vec}[1]{\mathbf{#1}}

\renewcommand{\phi}{\varphi}

\renewcommand{\rm}{\rrbracket}

\newcommand{\ev}[1]{\mathcal{F}_{#1}}
\newcommand{\glob}[1]{\mathcal{G}_{#1}}
\newcommand{\until}[1]{\mathcal{U}_{#1}}

\newcommand{\pstsp}{\bb{D}} 

	\title{A Robust Genetic Algorithm for Learning Temporal Specifications from Data }
	\author{Laura Nenzi\inst{1} \and 
    Simone Silvetti\inst{2,3} \and 
            Ezio Bartocci\inst{1} \and Luca Bortolussi\inst{4} }
	\authorrunning{L. Nenzi} 
	\tocauthor{Laura Nenzi}
	\institute{
        TU Wien, Vienna,  Austria
        \and
        DIMA, University of Udine, Udine, Italy
		\and
		Esteco S.p.A., Trieste, Italy
      \and 
        DMG, University of Trieste, Trieste, Italy
	}
	
\begin{document}

\maketitle
\thispagestyle{empty}
\pagestyle{empty}

\begin{abstract}
We consider the problem of mining signal temporal logical 
requirements from a dataset of regular (good) and anomalous 
(bad) trajectories of a dynamical system.  
We assume the training set to be labeled 
by human experts and that we have 
access only to a limited amount of data, typically noisy. 

We provide a systematic approach to synthesize both the 
syntactical structure and the parameters of the 
temporal logic formula using a two-steps procedure: first, we leverage a novel evolutionary algorithm for learning the structure of the formula; second, we perform the parameter synthesis operating on the statistical emulation of the average 
robustness for a candidate formula w.r.t. its parameters.

We compare  our results with our previous work~\cite{BufoBSBLB14}  and with a recently proposed decision-tree~\cite{bombara_decision_2016} based method. We present experimental results on two case studies: an anomalous trajectory detection problem of a naval surveillance system and the characterization of an Ineffective Respiratory effort, showing the usefulness of our work.
\end{abstract}

\addtolength{\textheight}{-1cm}   



\section{Introduction}
\label{sec:introduction}
Learning temporal logic requirements from data is an emergent 
research field gaining momentum in the rigorous engineering of 
cyber-physical systems.  
Classical machine learning methods typically generate 
very powerful black-box (statistical) models. However, 
these models often do not help in the comprehension of the 
phenomenon they capture.
Temporal logic provides a precise formal specification language 
that can easily be  interpreted by humans.  
The possibility to describe datasets in a concise way using temporal logic formulas can thus help to better clarify and comprehend which are the emergent patterns for 
the system at hand.  
A clearcut example is the problem of anomaly 
detection, where the input is a set of trajectories describing 
regular or anomalous behaviors, and the goal is to learn a 
classifier that not only can be used to detect anomalous 
behaviors at runtime, but also gives insights on what 
characterizes an anomalous behavior. 
Learning temporal properties is also relevant in combination 
with state of the art techniques for search-based falsification 
of complex closed-loop systems~\cite{0001SDKJ15,Sankaranarayanan17,BartocciBNS15,SilvettiPB17}, 
as it can provide an automatic way to describe desired 
(or  unwanted behaviors) that the system needs to satisfy. 

In this paper, we consider the problem of learning a temporal 
logic specification from a set of trajectories which 
are labeled by human experts (or by any other method 
which is not usable for real-time monitoring)
as ``good" for the normal expected behaviors and 
``bad" for the anomalous ones.
The goal is to automatically synthesize both 
the structure of the formula and its parameters 
providing a temporal logic classifier that 
can discriminate as much as possible the 
bad and the good behaviors. 
This specification can be turned into 
a monitor that output a positive verdict 
for good behaviors and a negative verdict
for bad ones. 

\paragraph*{\textbf{Related Work}}
 


Mining temporal logic requirements is an emerging field of research in the analysis 
of cyber-physical systems (CPS)~\cite{Xu2018,AckermannCHRSL10,AsarinDMN11,HoxhaDF18,BBS14,JinDDS15,nguyen_abnormal_2017,ZhouRWT17,kong_temporal_2017,bombara_decision_2016,BufoBSBLB14}. 
This approach is orthogonal to 
active automata learning (AAL) such as $L*$ Angluin's 
algorithm~\cite{Angluin87} and its recent variants~\cite{IsbernerHS14,SteffenHI12}.  
AAL is suitable to capture the behaviours of black-box reactive systems 
and it has been successfully  employed in the field of 
CPS to learn how to interact with the 
surrounding environments~\cite{Chen2013,FuTHC14}.  Mining temporal logic 
requirements has the following advantages with respect to AAL. 
The first is that it does not require to interact with a reactive system. 
AAL needs to query the system in order to learn a Mealy machine 
representing the relation between the input provided and the output observed.
Mining temporal logic requirements can 
be applied directly to a set of observed signals without 
the necessity to provide an input.
The second is the possibility to use temporal logic 
requirements within popular tools (such as Breach~\cite{Donze10} 
and S-TaLiRo~\cite{AnnpureddyLFS11}) for monitoring and falsification 
analysis of CPS models.

Most of the literature related to  temporal logic inference from data 
focuses in particular on the problem of learning the optimal 
parameters given a specific template formula~\cite{Xu2018,AsarinDMN11,HoxhaDF18,BBS14,JinDDS15,nguyen_abnormal_2017,ZhouRWT17}.  
In~\cite{AsarinDMN11}, Asarin et al. extend {\it the Signal Temporal Logic} (STL)~\cite{Maler2004}
with the possibility to express the time bounds of 
the temporal operators and the constants of the 
inequalities as parameters. They also
provide a geometric approach to identify the subset of the 
parameter space that makes a particular signal 
to satisfy an STL specification.  
 Xu et al. have recently proposed in~\cite{Xu2018}  a temporal 
logic framework called \emph{CensusSTL} 
for multi-agent systems that consists of 
\emph{an outer logic STL formula}  with a variable 
in the predicate representing the number of 
agents satisfying \emph{an inner 
logic STL formula}.  In the same paper the
authors propose also a new inference algorithm 
similar to~\cite{AsarinDMN11} that given the 
templates for both the \emph{inner} and \emph{outer}
formulas,  searches for the optimal parameter values
that make the two formulas capturing the trajectory data 
of a group of agents.
In~\cite{HoxhaDF18} the authors use the same parametric STL extension in 
combination with the quantitative semantics of STL to
perform a counter-example guided inductive parameter 
synthesis.  This approach consists in iteratively 
generating a counterexample by executing a 
falsification tool for a template formula.
The counterexample found at each step is then 
used to further refine the parameter set and 
the procedure terminates when no other 
counterexamples are found.
In general, all these methods, when working directly 
with raw data, are potentially vulnerable to the
noise of the measurements and they are limited 
by the amount of data available.  

Learning both the structure and the parameters of a 
formula from a dataset poses even more challenges~\cite{kong_temporal_2017,bombara_decision_2016,BBS14,BufoBSBLB14}.
This problem is usually addressed in two steps, 
learning the structure of the formula and synthesizing its parameters.
In particular, in~\cite{kong_temporal_2017} the structure 
of the formula is learned by exploring a directed acyclic graph and the method  exploits {\it Support Vector Machine} (SVM) 
for the parameter optimization.  In~\cite{bombara_decision_2016} the authors use instead a \emph{decision tree} based approach for learning the formula, while the optimality is evaluated using heuristic 
impurity measures.

In our previous works~\cite{BBS14,BufoBSBLB14} we have also
addressed the problem of learning both the structure and the
parameters  of a temporal logic specification from data.  In~\cite{BBS14} the structure 
of the formula is learned using a heuristics algorithm, while  
in~\cite{BufoBSBLB14} using a genetic algorithm. The 
synthesis of the parameters is instead performed in both cases 
exploiting the {\it Gaussian Process Upper Confidence Bound} 
(GP-UCB)~\cite{gpucb} algorithm, statistically emulating
the satisfaction probability of a formula for a given set of parameters. 
In both these methodologies, it is required to learn first a statistical 
model from the training set of trajectories. However, the statistical 
learning of this model can be a very difficult task and this is one of 
the main reason for proposing our new approach.


\paragraph*{\textbf{Our contribution}}


In this work, we consider the problem of mining the formula directly from a data set without requiring to learn a generative model from data.
To achieve this goal,  we introduce a number of techniques to improve the potentials of the genetic algorithm and to deal with the noise in the data in the absence of an underlying model. 

First, instead of using the probability satisfaction as evaluator for the best formula, we design a discrimination function based on the quantitative semantics (or robustness) of STL and in particular the average robustness introduced in~\cite{BartocciBNS15}. 
The average robustness enables us to differentiate among STL classifiers that have similar discriminating performance with respect to the data by choosing the most robust one.  This gives us more information than just having the probability of satisfaction: for each trajectory, we can evaluate how much is it closed to the  ``boundary''  of the classifier, instead of only checking whether it satisfies or not a formula. 
We then modify the discrimination function and the GP-UCB algorithm used in~\cite{BBS14} and~\cite{BufoBSBLB14} to better deal with noisy data and to use the quantitative semantics to emulate the average robustness distribution with respect to the parameter space of the formula.

Second, we reduce the computational cost of the \emph{Evolutionary Algorithm} (EA) by using a lightweight configuration (i.e., a low threshold of max number of iterations) of the GP-UCB optimization algorithm to estimate the parameters of the formulas at each generation.  
The EA algorithm generates, as a final result, an STL formula tailored for classification purpose. 

We compare our framework with our previous methodology 
\cite{BufoBSBLB14} and with the decision-tree based approach presented in~\cite{bombara_decision_2016} on an anomalous trajectory detection problem of a naval surveillance and on an Assisted Ventilation in Intensive Care Patients. 
Our experiments indicate that the proposed approach 
outperforms our previous work with respect to accuracy and show that it produces in general more compact, and easy to read, temporal logic specifications with respect to the decision-tree based  approach with a comparable speed and accuracy.

\paragraph*{\textbf{Paper structure}} 
The rest of the paper is organized as follows: in the next section we present the {\it Signal Temporal Logic} and its robust semantics. We then introduce the problem considered in Section  \ref{sec:problem}. In section \ref{sec:methodology}, we describe our approach. The results are presented in Section \ref{sec:results}. Finally, we conclude the paper in Section \ref{sec:conclusion}, by discussing the implications of our contribution, both from the practical and the methodological aspect and some directions of improvement.
 
\section{Signal Temporal Logic}
\label{sec:background}

\paragraph*{ \bf STL} {\it Signal Temporal Logic} (STL)~\cite{Maler2004} is a formal 
specification language to express temporal properties over 
real-values trajectories with dense-time interval. For the rest of the paper, let be $\vec x: \mathbb{T} \rightarrow \mathbb{R}^n$ a trace/trajectory, where $ \mathbb{T}  = \mathbb{R}_{\geq 0}$ is the time domain, $x_i(t)$ is the value at time $t$ of the projection on the $i^{th}$ coordinate, and $\vec x = (x_1, ..., x_n)$, as an abuse on the notation, is used also to indicate the set of variables of the trace considered in the formulae.
\begin{definition} [STL syntax] The syntax of STL is given by
$$\varphi :=  
    \top \:|\: \mu  \:|\:  \neg \varphi \:|\: \varphi_{1} \wedge \varphi_{2}  \:|\:  \varphi_{1} \: \until{I} \: \varphi_{2}  $$
where  $\top$ is the Boolean true constant, $\mu$ is an {\it atomic proposition}, inequality of the form $y(x) > 0$ ($y:\mathbb{R}^n \rightarrow \mathbb{R}$), {\it negation} $\neg$ and {\it conjunction} $\wedge$ are the standard Boolean connectives, and $\until{I}$ is the {\it Until}  temporal modality, where  $I$ is a real positive interval.
As customary, we can derive the {\it disjunction} operator $\vee$ and the future  {\it eventually} $\ev{I}$ and  {\it always} $\glob{I}$ operators from  the until temporal modality $( \varphi_1 \vee \varphi_2 = \neg (\neg \varphi_1 \wedge \neg \varphi_2),$ $\ev{I} \varphi = \top \until{I} \varphi,\mbox{ } \glob{I} \varphi = \neg  \ev{I} \neg \varphi ).$
\end{definition}

 STL can be interpreted over a trajectory $\vec x$ using a 
 qualitative (yes/no) or a quantitative (real-value) semantics~\cite{Maler2004,Donze2013}.  
 We report here only the quantitative semantics and we refer the reader 
 to~\cite{MalerN13,Maler2004,Donze2013} for more details.
\begin{definition} [{STL  Quantitative Semantics}]
\label{sp.rob}
The \emph{quantitative satisfaction function} $\rho$ returns a value 
$\rho(\varphi, \vec{x},t) \in \bar{\mathbb{R}}$,\footnote{$\bar{\mathbb{R}}= \mathbb{R}\cup \{ -\infty, +\infty\}.$} 
quantifying the robustness degree (or satisfaction degree) of the property $\varphi$ by the trajectory $\vec{x}$ 
at time $t$.  It is defined recursively as follows:
{\small
\begin{align*}
&\rho(\top, \vec x, t)  & = &\mbox{ } \phantom{a}  {\large + \infty} \\
&\rho(\mu, \vec{x},t)  & = &\mbox{ } \phantom{a}  y(\vec{x}(t))  \mbox{   where } \mu \equiv y(\vec{x}(t)) \geq 0 \\ 
& \rho (\neg \varphi,\vec{x},t)  & =  & \mbox{ } \phantom{a}  - \rho (\varphi,\vec{x},t)\\
&\rho( \varphi_{1} \wedge  \varphi_{2}, \vec{x},t)  & = &\mbox{ } \phantom{a} \min ( \rho( \varphi_{1},\vec{x},t),\rho( \varphi_{2},\vec{x},t) ) \\
& \rho(  \varphi_{1} \: \until{[a,b]}  \varphi_{2}, \vec{x},t) & = &\mbox{ } \phantom{a}  \sup_{t'\in t+[a,b]}(\min(\rho( \varphi_{2},\vec{x},t'),\inf_{t'' \in [t,t')}(\rho( \varphi_{1},\vec{x},t''))))
\end{align*}
}
Moreover, we let $\rho(\varphi, \vec{x}):=\rho(\varphi, \vec{x},0)$.
\end{definition}
The sign of $\rho(\varphi, \vec{x})$ provides the link with the standard Boolean semantics 
of~\cite{Maler2004}: $\rho(\varphi, \vec{x})>0$  only if  $\vec{x}\models \varphi$, while $\rho(\varphi, \vec{x})<0$ 
only if $\vec{x}\not\models\varphi$\footnote{The case  $\rho(\varphi, \vec{x}) = 0$, instead, is a borderline case, and 
the truth of $\varphi$ cannot be assessed from the robustness degree alone.}.
The absolute value of $\rho(\varphi, \vec{x})$, instead,  can be interpreted as a measure of the robustness of the satisfaction with respect to noise in signal $\vec{x}$, measured in terms of the maximal
perturbation in the secondary signal $  y(\vec{x}(t)) $, preserving truth value. This means that if $\rho(\varphi,\vec{x},t)=r$  then for every signal 
$\vec{x}'$ for which every secondary signal satisfies $\max_{t}| y_{j}(t) - y'_{j}(t)|<r$, we have that 
$\vec{x}(t) \models \varphi$ if and only if $\vec{x'}(t)\models\varphi$ ({\it correctness property}).

\paragraph*{\bf PSTL} {\it Parametric Signal Temporal Logic}~\cite{AsarinDMN11} is an extension 
of STL that parametrizes the formulas. There are two types of formula parameters: 
temporal parameters, corresponding to the time bounds in the time intervals associated 
with temporal operators (e.g. $a,b \in \mathbb{R}_{\geq 0}$, with $a<b$, s.t. $\ev{[a,b]} \mu$), 
and the threshold parameters, corresponding to the constants in the inequality predicates 
(e.g., $k \in \mathbb{R}, \mu = x_i > k$, where $x_i$ is a variable of the trajectory). 
In this paper, we allow only atomic propositions of the form $\mu = x_i \bowtie k$ with
 $\bowtie \in \{ >, \leq \}$. 
Given an STL formula $\phi$, let  $ \mathbb{K}=(\mathcal{T}\times\mathcal{C})$ be
the {\it parameter space}, where $\mathcal{T} \subseteq \mathbb{R}_{\geq 0}^{n_t}$ 
is the temporal parameter space ($n_t$ being the number of temporal parameters), 
and $\mathcal{C}   \subseteq  \mathbb{R}^{n_k}$ is the threshold parameter space 
($n_k$ being the number of threshold parameters). A $\boldsymbol{\theta} \in \mathbb{K}$ 
is a parameter configuration that induces a corresponding STL formula; e.g., 
$\phi=\ev{[a,b]}( x_i > k), \boldsymbol{\theta} = (0,2,3.5)$ then $\phi_{ \boldsymbol{\theta}}=\ev{[0,2]}( x_i > 3.5)$.

\paragraph*{\bf  Stochastic Robustness} Let us consider an unknown stochastic process $({\bf X}(t))_{t \in T}=(X_{1}(t),...,X_{n}(t))_{t \in T}$, where each vector ${\vec X}(t)$ corresponds to the state of the system at time $t$.  
For simplicity, we indicate the stochastic model with $\vec X(t)$.  
$\vec{X}(t)$ is a measurable also as a random variable $\vec{X}$ on the space $\sspace{}$-valued {\it cadlag functions} ${\cal{D}}([0,\infty),\sspace{})$, here denoted by ${\cal{D}}$, assuming the domain $\pstsp$ to be fixed. It means that the set of  trajectories $\vec x$  of the stochastic process $\vec{X}$ is the set ${\cal{D}}$. Let us consider now an STL formula $\phi$, with predicates interpreted over state variables of ${\vec X}$. Given a trajectory $\vec{x}(t)$ of a stochastic system,  its robustness $\rho(\phi,\vec{x},0)$ is a measurable functional $R_\phi$~\cite{BartocciBNS15} from the trajectories in $\cal{D}$ to $\mathbb{R}$  
which defines the real-valued random variable $R_{\phi} = R_{\phi}(\vec{X})$ with probability distribution:
 \[\mathbb{P}\left(R_{\phi}(\vec{X})\in [a,b] \right)= \mathbb{P}\left( \vec{X} \in \{\vec{x}\in\mathcal{D}~|~\rho(\phi,\vec{x},0)\in [a,b] \}   \right).  \]
Such distribution of robustness degrees can be summarized by the average robustness degree. Fixing the stochastic process $\vec X$, $\mathbb{E}(R_{\phi}| \vec{X})$,  it gives a measure of how strongly a formula is satisfied. 
The satisfaction is more robust when this value is higher. In this paper, we will approximate this expectation by Monte Carlo sampling.

 
\section{Problem formulation}
\label{sec:problem}

In this paper, we focus our attention on learning the best property 
(or set of properties) that  discriminates trajectories belonging to 
two different classes, say good and bad, starting from a labeled 
dataset of observed trajectories. Essentially, we want to tackle 
a supervised two-class classification problem over trajectories, 
by learning a temporal logic discriminant, describing the temporal 
patterns that better separate two sets of observed trajectories.


The idea behind this approach is that there exists an unknown 
procedure that, given a temporal trajectory, is able to decide if 
the signal is good or bad. This procedure can correspond to 
many different things, e.g., to the reason of the failure of 
a sensor that breaks when it receives certain inputs.
In general, as there may not be an STL specification
 that perfectly explains/mimics the unknown procedure,  
our task is to approximate it with the most effective one.

Our approach works directly with observed data, and 
avoids the reconstruction of an intermediate generative 
model $p(\vec x|z)$ of trajectories $x$ conditioned on 
their class $z$, as in \cite{BufoBSBLB14,BBS14}. 
The reason is that such models can be hard to construct, 
even if they provide a powerful regularization, as they 
enable the generation of an arbitrary number of samples 
to train the logic classifier. 

In a purely data-driven setting, to build an effective classifier, 
we need to consider that training data is available in limited 
amounts and it is typically noisy.  This reflects in the necessity 
of finding methods that guarantee good generalization 
performance and avoid overfitting. 
In our context, overfitting can result in overly complex formulae, 
de facto encoding the training set itself rather than guessing 
the underlying patterns that separate the trajectories.  This can 
be faced by using a score function based on robustness of 
temporal properties, combined with suitably designed 
regularizing terms. 

We want to stress that the approach we present here, 
due to the use of the average robustness of STL properties, 
can be easily tailored to different problems,  like finding the 
property that best characterise a single set of observations.


 
\section{Methodology}
\label{sec:methodology}

Learning an STL formula can be separated in two optimization problems: the learning of the formula structure and the synthesis of the formula parameters. 
The structural learning is treated as a discrete optimization problem using an {\it Genetic Algorithm} (GA); the parameter synthesis, instead, considers a continuous parameter space and exploits an active learning algorithm, called {\it Gaussian Process Upper Confidence Bound (GP-UCB)}.
The two techniques are not used separately but combined together in a bi-level optimization. The GA acts externally by defining a set of template formulae. Then, the GP-UCB algorithm, which acts at the inner level, finds the best parameter configuration such that each template formula better discriminates between the two datasets. 
For both optimizations, we need to define a score function to optimize, encoding the criterion to discriminate between the two datasets.

\begin{algorithm}[htbp] 
\caption{ROGE -- RObustness GEnetic }
\label {algo:ROGE}
\vspace{1mm}
\begin{algorithmic}[1]
\REQUIRE $\mathcal{D}_p, \mathcal{D}_n, \mathbb{K}, Ne, Ng, \alpha, s$
\STATE $gen \gets \textsc{generateInitialFormulae}(Ne,s)$
\STATE $gen_{\Theta} \gets \textsc{learningParameters}(gen, G, \mathbb{K})$
\FOR {$i=1 \dots Ng$}
	\STATE $subg_{\Theta} \gets \textsc{sample}(gen_{\Theta} ,F)$
	\STATE $newg \gets \textsc{evolve}(subg_{\Theta},\alpha)$
    \STATE $newg_{\Theta} \gets \textsc{learningParameters}(newg, G, \mathbb{K})$
	\STATE $gen_{\Theta} \gets \textsc{sample}(newg_{\Theta} \cup gen_{\Theta}, F)$
\ENDFOR
\RETURN $gen_{\Theta}$
\end{algorithmic}
\end{algorithm}

Our implementation, called \emph{RObustness GEnetic} (ROGE) algorithm is described in  Algorithm \ref{algo:ROGE}.  First, we give an overview of it and then we described each specific function in the next subsections. 
The algorithm requires as input the dataset $\mathcal{D}_p$(good) and $\mathcal{D}_n$(bad), the parameter space $\mathbb{K}$, with the bound of each considered variable, the size of the initial set of formulae $Ne$, the number of maximum iterations $Ng$, the mutation probability $\alpha$ and the maximum initial formula size $s$.   
The algorithm starts generating (line 1) an initial set of PSTL formulae, called $gen$. 
For each of these formulae (line 2), the algorithm learns the best parameters accordingly to a discrimination function $G$.  We call $gen_{\Theta}$ the generation for which we know the best parameters of each formula.
During each iteration, the algorithm (line 4), guided by a fitness function F , extracts  a subset $subg_{\Theta}$ composed by the best $Ne/2$ formulae, from the initial set $gen_{\Theta}$.
From this subset (line 5), a new set $newg$ composed of $Ne$ formulae is created by employing the \textsc{Evolve} routine, which implements two {\it genetic} operators. 
Then (line 6), as in line 2, the algorithm identifies the best parameters of each formula belonging to $newg$. 
The new generation $newg_{\Theta}$ and the old generation $gen_{\Theta}$ are then merged together (line 7). From this set of $2Ne$ formulae the algorithm extracts, with respect to the fitness function $F$, the new generation $gen_{\Theta}$ of $Ne$ formulae. 
At the end of the iterations or when the  $\textsc{stop}$ criterion is true (lines 8-12), the algorithm returns the last generated formulae. 
The best formula is the one with the highest value of the discrimination function, i.e., $\phi_{best} =  \mathrm{arg max}_{\phi_{ \boldsymbol{\theta}} \in gen_{\Theta}}(G(\phi_{\boldsymbol{\theta}}))$.
We describe below in order: the discrimination function algorithm, the learning of the parameters of a formula template and the learning of the formula structure.

\subsection{\bf Discrimination Function $G(\phi)$}
We have two datasets $\mathcal{D}_p$ and $\mathcal{D}_n$ and we search for the formula $\phi$  that better separates these two classes. We define a function able to discriminate between this two datasets, such that maximising this \emph{discrimination function} corresponds to finding the best formula classifier. 
In this paper, we decide to use, as evaluation of satisfaction of each formula, the quantitative semantics of STL. 
As described in Section~\ref{sec:background}, this semantics computes a real-value (robustness degree) instead of only a yes/no answer. 

Given a  dataset $\mathcal{D}_i$, we  assume that the data comes from an unknown stochastic process $\vec X_i$. The process in this case is like a black-box for which we observe only a subset of trajectories, the dataset $\mathcal{D}_i$.
We can then estimate the averages robustness $\bb{E}(R_{\phi}|\vec X_i) $ and its variance $\sigma^2(R_{\phi} | \vec X_i)$, averaging over $\mathcal{D}_i$.

To discriminate between the two dataset $\mathcal{D}_p$ and $\mathcal{D}_n$, we search the formula $\phi$ that maximizes the difference between the average robustness of  $\vec X_p$, $\bb{E}(R_{\phi} | \vec X_p) $, and the average robustness of  $\vec X_n$, $\bb{E}(R_{\phi}| \vec X_n) $ divided by the sum of the respective standard deviation:
\begin{equation}
\label{eqn:discrfcn}
G({\phi}) =  \frac{\bb{E}(R_{\phi} | \vec X_p) - \bb{E}(R_{\phi} | \vec X_n)}{\sigma(R_{\phi} | \vec X_p) + \sigma(R_{\phi} | \vec X_n) }.
\end{equation}
This formula is proportional to the probability that a new point sampled from the distribution generating $\vec X_p$ or  $\vec X_n$, will belong to one set and not to the other. In fact, an higher value of $G({\phi})$ implies that the two average robustness will be sufficiently distant relative to their  length-scale, given by the standard deviation $\sigma$.


As said above, we can  evaluate only a statistical approximation of $G({\phi})$ because we have a limited number of samples belonging to $\vec X_p$ and  $\vec X_n$.
We will see in the next paragraph how we tackle this problem.

\subsection{\bf GB-UCP: learning the parameters of the formula}
\label{sec:methodology:GPUCB}
Given a formula $\phi$ and a parameter space  $ \mathbb{K}$, we want to find the parameter configuration $\boldsymbol{\theta} \in  \mathbb{K}$ that maximises the score function $g (\boldsymbol{\theta}) := G(\phi_{\boldsymbol{\theta}})$, considering that the evaluations of this function are noisy. So, the question is how to best estimate (and optimize) an unknown function from observations of its value at a finite set of input points. This is a classic non-linear non-convex optimization problem that we tackle by means of 
the GP-UCB algorithm \cite{gpucb}. 
This algorithm interpolates the noisy observations using a stochastic process 
(a procedure called emulation in statistics) and optimize the emulated function using the uncertainty of the fit to determine regions where the true maximum can lie. 
More specifically, the GP-UCB  bases its emulation phase  on Gaussian Processes, a Bayesian non-parametric regression approach~\cite{Rasmussen2007}, adding candidate maximum points to the training set of the GP in an iterative fashion, and terminating when no improvement is possible (see \cite{gpucb} for more details).

After this optimization, we have found a formula that separates the two datasets, not from the point of view of the satisfaction (yes/no) of the formula but from the point of view of the robustness value. In other words, there is a threshold value $\alpha$ such that  $\bb{E}(R_{\phi}|\vec X_p) > \alpha$ and $\bb{E}(R_{\phi}|\vec X_n) \leq \alpha$. Now, we consider the new STL formula $\phi'$ obtained translating the atomic predicates of $\phi$ by $\alpha$ (e.g., $y(x)>0$ becomes $y(x)-\alpha>0$). Taking into account that the quantitative robustness is achieved by combination of $\min$, $\max$, $\inf$ and $\sup$, which are linear algebraic operators with respect to translations (e.g, $\min(x,y)\pm c = \min(x \pm c,y \pm c)$), we easily obtain that $\bb{E}(R_{\phi'}|\vec X_p) > 0$ and $\bb{E}(R_{\phi'}|\vec X_b)< 0$. Therefore, $\phi'$ will divide the two datasets also from the point of view of the satisfaction. 


\subsection{\bf Genetic Algorithm: learning the structure of the formula} 
 To learn the formula structure, we exploit a modified version of the Genetic Algorithm (GA) presented in  \cite{BufoBSBLB14}. GAs belongs to the larger class of evolutionary algorithms (EA). They are used for search and optimization problems. The strategy takes inspiration from the genetic area, in particular in the selection strategy of species.  Let us see now in detail the genetic routines of the ROGE algorithm.

\paragraph*{\bf $gen = \textsc{generateInitialFormulae}(Ne,s)$.} This routine generates the initial set of $Ne$ formulae. A fraction $n_l < Ne$ of this initial set is constructed by a subset of the temporal properties: $\ev{I}\mu$,  $\glob{I}\mu$, $\mu_1\until{I}\mu_2$, where the atomic predicates are of the form $\mu = (x_i > k)$  or $\mu = (x_i \le k)$ or simple boolean combinations of them (e.g. $\mu = (x_i > k_i) \wedge (x_j > k_j)$). The size of this initial set is exponential accordingly to the number of considered variables $x_i$. If we have few variables we can keep all the generated formulae, whereas if the number of variables is large we consider only a random subset. The remain $n_r = Ne - n_l$ formulae are chosen randomly from the set of formulae with a maximum size defined by the input parameter $s$. This size can be adapted to have a wider range of initial formulae.

\paragraph*{\bf $ subg_{\Theta} = \textsc{sample}(gen_{\Theta} ,F)$.}
This procedure extracts from $gen_{\Theta}$ a subset $subg_{\Theta}$ of $Ne/2$ formulae, according to a fitness function $F(\phi) = G(\phi) - S(\phi)$.
The first factor $G(\phi)$ is the dicrimination function previously defined and $S(\phi)$ is a size penalty, i.e. a function penalizes formulae with large sizes.  In this paper, we consider $S(\phi) = g \cdot p^{size(\phi)}$, where $p$ is heuristically set such that $p^5 = 0.5$, i.e. formulae of size 5 get a 50\% penalty, and $g$ is adaptively computed as the average discrimination in the current generation. An alternative choice of $p$ can be done by cross-validation.

\paragraph*{\bf $ newg = \textsc{evolve}(subg_{\Theta},\alpha)$.}
This routine defines a new set of formulae ($newg$)  starting from $subg_{\Theta}$, exploiting  two {\it genetic} operators: the {\it recombination}  and the {\it mutation} operator.
The recombination operator takes as input two formulae $\phi_1$, $\phi_2$ (the parents), it randomly chooses a subtree from each formula and it swaps them, i.e. it assigns the subtree of $\phi_1$ to $\phi_2$ and viceversa. As a result, the two generated formulae (the children) share different subtrees of the parents' formulae. 
The mutation operator takes as input one formula and induce a random change (such as inequality flip, temporal operator substitution, etc.) on a randomly selected node of its tree-structure. The purpose of the genetic operators is to introduce innovation in the offspring population which leads to the optimization of a target function ($G$ in this case). In particular,  recombination is related to exploitation, whereas  mutation to exploration.
The \textsc{evolve} routine implements an iterative loop that at each iteration  selects which genetic operators to apply. A number $p \in [0,1]$ is randomly sampled.  If its value is lower than the mutation probability, i.e.,  $p \le \alpha$, then the mutation is selected, otherwise a recombination is performed.
At this point the input formulae of the selected genetic operator are chosen randomly between the formulas composing $subg_{\Theta}$ and the genetic operators are applied.
In our implementation, we give more importance to the exploitation; therefore the mutation acts less frequently than the recombination (i.e., $\alpha \le 0.1$).
The iteration loops will stop when the number of generated formula is equal to $Ne$, i.e. twice the cardinality of $subg_\Theta$.

\section{Case studies and Experimental Results}
\label{sec:results}

\subsection{Maritime Surveillance}
As first case study, We consider the maritime surveillance problem presented in~\cite{bombara_decision_2016} to compare our framework with their {\it Decision Tree} (DTL4STL) approach.  The experiments with the DTL4STL approach were implemented 
in \texttt{Matlab}, the code is available at~\cite{DTL4STL}.
We also test our previous implementation presented 
in~\cite{BufoBSBLB14} with the same benchmark. 
Both the new an the previous learning procedure 
were implemented in \texttt{Java} (JDK 1.8\_0) and run on a Dell XPS, Windows 10 Pro, Intel Core i7-7700HQ 2.8 GHz, 
8GB 1600 MHz memory.

\begin{figure}[thpb]
      \centering
       \includegraphics[width=1.1 \textwidth]{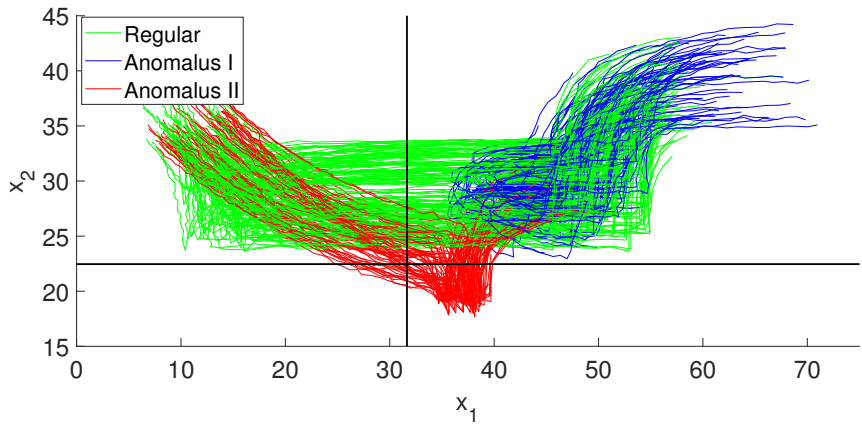}
      \caption{The regular trajectories are represented in green. The anomalous trajectories which are of two kinds, are represented respectively in blue and red.}
      \label{fig:naval dataset}
   \end{figure}

The synthetic dataset of naval surveillance reported 
in~\cite{bombara_decision_2016} consists of 2-dimensional 
coordinates traces of vessels behaviours.  It considers two kind 
of anomalous trajectories and regular trajectories, as illustrated in Fig.~\ref{fig:naval dataset}.  The dataset contains 2000 total trajectories (1000 normal and 1000 anomalous) with 61 sample points per trace.  
We run the experiments using a 10-fold cross-validation in order to collect the mean and variance of the misclassified trajectories of the validation set. 
Results are summarized in Table~\ref{tab:navalResults}, where we report also the average execution time.
We test also our previous implementation~\cite{BufoBSBLB14}
without a generative model from data. It performs so 
poorly on the chosen benchmark that is not meaningful to report it: the misclassification rate on the validation set is around 50\%.
In Table \ref{tab:navalResults}, we also report DTL4STL$_p$ the DTL4STL performance declared in~\cite{bombara_decision_2016}, but we were not be able to reproduce them in our setting.

\begin{table}[!t]
\begin{center}
\scalebox{1.15}{
\begin{tabular}{|c|c|c|c|c|c|c|}
\hline
& ROGE &DTL4STL & DTL4STL$_p$ \\
\hline
Mis. Rate &$0$  &  $0.01 \pm 0.013$& $0.007\pm0.008$ \\
\hline
Comp. Time &$73 \pm18$ & $144 \pm24$& -  \\

\hline
\end{tabular}}
\end{center}
\caption{The comparison of the computational time (in sec.), the mean missclassification rate and its standard deviation between the learning procedure using the RObust GEnetic algorithm, the Decision-Tree (DTL4STL) Matlab code provided by the authors and the results reported in  \cite{bombara_decision_2016} (DTL4STL$_p$).}
\label{tab:navalResults}
\end{table}
In terms of accuracy our approach is comparable with respect to the performance of the DTL4STL. In terms of computational cost, the decision tree approach is slightly slower but it is implemented in \texttt{Matlab} rather than \texttt{Java}.

An example of formula that we learn with ROGE is
\begin{equation}
\label{eqn:bestourformula}
\phi=((x_2 > 22.46)\, \mathcal{U}_{[49, 287]}\, (x_1 \le 31.65))
\end{equation}
The formula (\ref{eqn:bestourformula}) identifies all the regular trajectories, remarkably resulting in a misclassification error equal to zero, as reported in Table \ref{tab:navalResults}. The red anomalous trajectories falsify the predicate $x_2 > 22.46$ before verifying $x_1\le31.65$, on the contrary the blue anomalous trajectories globally satisfy $x_2 > 22.46$ but never verify $x_1 \le 31.65$ (consider that all the vessels start from the top right part of the graphic in Figure \ref{fig:naval dataset}). Both these conditions result in the falsification of Formula(~\ref{eqn:bestourformula}), which on the contrary is satisfied by all the regular trajectories. In Figure \ref{fig:naval dataset}, we have reported the threshold $x_2=22.46$ and $x_1=31.65$. 
 

An example instead of formula found by the DTL4STL tool using the same dataset is the following:
\begin{eqnarray*}
\small
\label{eqn:bestDTformula}
\psi =
(((\mathcal{G}_{[187,196)}x_{1}<19.8) \wedge
(\mathcal{F}_{[55.3,298)}x_{1}>40.8) ) \vee ((\mathcal{F}_{[187,196)}x_{1}>19.8) \wedge \\ ((\mathcal{G}_{[94.9,296)}x_{2}<32.2)  \vee ((\mathcal{F}_{[94.9,296)}x_{2}>32.2) \wedge (((\mathcal{G}_{[50.2,274)}x_{2}>29.6) \wedge\\
  (\mathcal{G}_{[125,222)}x_{1}<47) ) \vee
   ((\mathcal{F}_{[50.2,274)}x_{2}<29.6) \wedge (\mathcal{G}_{[206,233)}x_{1}<16.7) )))))
\end{eqnarray*}

The specific formula generated using ROGE is simpler than the formula generated using DTL4STL and indeed it is easier to understand it. Furthermore DTL4STL does not consider the until operator in the set of primitives (see \cite{bombara_decision_2016} for more details).

\subsection{Ineffective Inspiratory Effort (IIE)}
\begin{figure}[thpb]
      \centering
      \includegraphics[width=1.\textwidth]{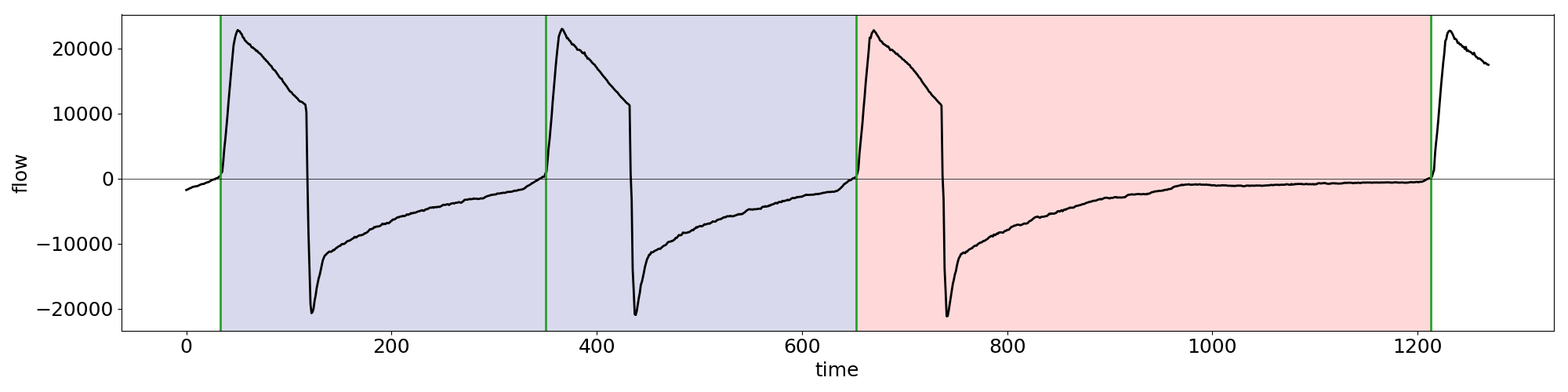}  
      \caption{Example of regular (blue regions) and ineffective (red region) breaths.}
      \label{fig:breaths}
  \end{figure}

The Ineffective Inspiratory Effort (IIE) is one of the major problems concerning the mechanical ventilation of patients in intensive care suffering from respiratory failure.
The mechanical ventilation uses a pump to help the patient in the process of ispiration and expiration, controlling the $\mathit{flow}$ 
of air into the lugs and the airway pressure ($paw$).
Inspiration is characterised by growth of pressure up to the selected $paw$ value and positive $\mathit{flow}$, while expiration has a negative $\mathit{flow}$ and a drop $paw$. The exact dynamics of these  respiratory curves stricly depends on patient and on ventilatory modes. We can see an example in~Fig.~\ref{fig:breaths} of two regular (blue regions) and one ineffective (red region) breaths. An IIE occurs when the patients tries to inspire, but its effort is not sufficient to trigger the appropriate reaction of the ventilator. This results in a longer expiration phase. Persistence of IIE may cause permanent damages of the respiratory system.

An early detection of anomalies using low-cost methods is a key challenge to prevent IIE conditions and still an open problem.
In~\cite{BufoBSBLB14}, starting with a dataset of discrete time series and sampled $\mathit{flow}$ values (labeled good and bad) from a single patient, the authors statically designed generative models of $\mathit{flow}$ and of its derivative $\mathit{flow}'$, for regular and ineffective signals. Then they used the simulations of such models to identify the best formula/formulae discriminating between them. In that contribution trajectories with different lengths are considered, treating as false trajectories that are too short to verify a given formula, and use this to detect the length of trajectories and separate anomalous breaths which last longer than regular ones.  
However, using the information about the duration of a breath is not advisable if the goal is to monitor the signals at runtime and detect the IEE at their onset. 
For this reason, in our approach we consider a new dataset, cutting the original trajectories used to generate the stochastic model in Bufo et al.~\cite{BufoBSBLB14} to a maximum common length of the order of 2 seconds. We also apply a moving average filter to reduce the noise in the computation of $\mathit{flow}'$. 


Specifically, the new dataset consists of 2-dimensional traces of $\mathit{flow}$ and its derivative, $\mathit{flow}'$, containing a total of 288 trajectories (144 normal and 144 anomalous) with 134 sample points per trace. The time scale is in hundredths of a second.
We run the experiments using a 6-fold cross-validation and report our results and comparison on the new dataset with DTL4STL~\cite{bombara_decision_2016} in Table~\ref{tab:breathResults}. 


\begin{table}[!t]
\begin{center}
\scalebox{1.15}{
\begin{tabular}{|c|c|c|c|c|c|c|}
\hline
& ROGE & DTL4STL \\
\hline
Mis. Rate & $0.17\pm 0.01$  & $0.23\pm0.07$  \\
\hline
False. Pos & $0.20\pm 0.02$  & $0.23\pm0.07$ \\
\hline
False. Neg & $0.14\pm 0.02$  & $0.20\pm0.15$ \\
\hline
Comp. Time & $65\pm7$ & $201 \pm 7$  \\
\hline
\end{tabular}}
\end{center}
\caption{The comparison of the computational time (in sec.), the mean missclassification rate and its standard deviation between the learning procedure using the RObust GEnetic algorithm, }
\label{tab:breathResults}
\end{table}


An example of formula that we learn with ROGE is:
\begin{equation}
\label{eqn:bestourformulaIIE1}
\phi =  \ev{[31, 130]}( (\mathit{flow} \ge -670) \vee (\mathit{flow}' \le -94))
\end{equation}



\begin{figure}[thpb]
      \centering      \includegraphics[width=1.\textwidth]{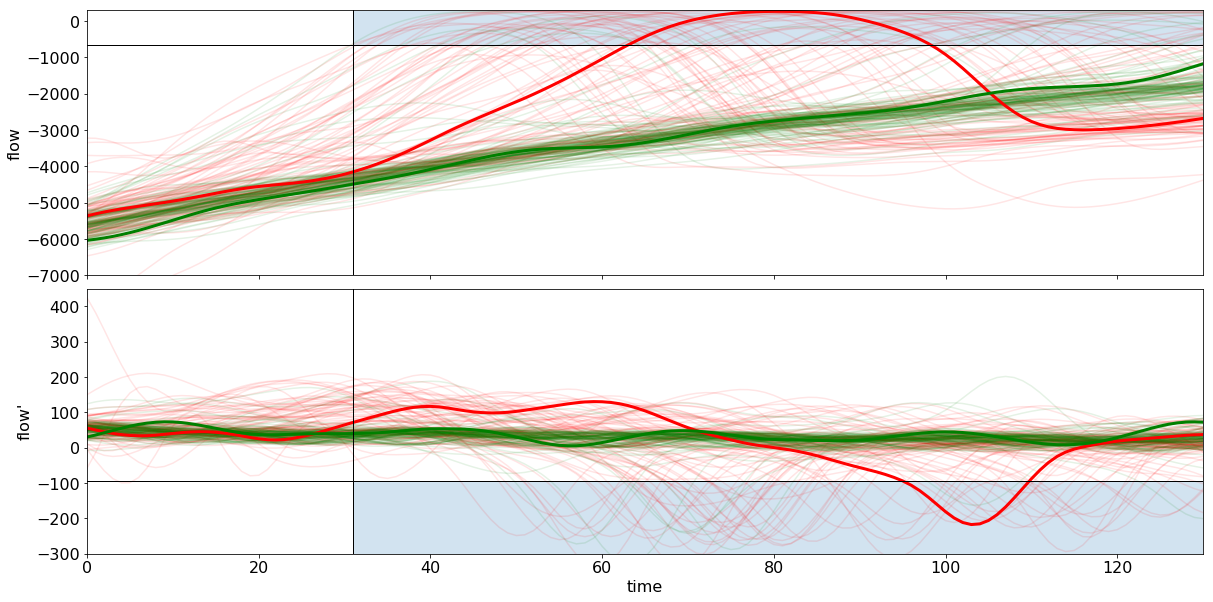} 
      \caption{  100 regular (green) and  100 ineffective (red) flow  and $\mathit{flow}'$ trajectories during the expiration phase. The light blue rectangles correspond to the satisfaction area of formula~(\ref{eqn:bestourformulaIIE1}). One regular (green) and one ineffective (red) trajectories are showed thicker.}
      \label{fig:hump}
   \end{figure}

Formula $\phi$ identifies the anomalous trajectories, stating that at a time between $31$ sec and $130$ seconds either $\mathit{flow}$ is higher than $-670$ or  $\mathit{flow}'$ is below $-94$.
This is in accordance with the informal description of an IIE, principally caused by an unexpected increment of the $\mathit{flow}$ during the expiration phase followed by a rapid decrease. Indeed, one of the main characteristic of an IIE is the presence of a small hump in the $\mathit{flow}$ curve during this phase. In Fig.~\ref{fig:hump} we report some of the trajectories of the dataset, showing the areas where the property is satisfied.

On average, formulae found by the DTL4STL tool on the new dataset are disjunctions or conjunctions of$~10$ eventually or always subformulae, which are not readily interpretable. 

Our approach on the new dataset is better in term of accuracy and computation time with an improvement of 22\% and 67\%, respectively.

Similarly to the previous test case, we compare our approach with \cite{BufoBSBLB14}, performed directly on the dataset, and not on the generative model. That approach performs so poorly also in this benchmark, obtaining a misclassification rate of $0.47$, which is comparable with a random classifier. The problem here is that the methods proposed in \cite{BufoBSBLB14}, differently from the one presented here, relies on a large number of model simulations, and it is not suited to work directly with observed data.

If we give up to online monitoring and consider only full breaths, we can improve classification by rescaling their duration into [0,1], so that each breath lasts the same amount of time. In this case, we learn a formula with misclassification rate of $0.05\pm0.01$, while DTL4STL reaches a misclassification rate of $0.07\pm0.02$. This suggests that the high variability of durations of ineffective breaths has to be treated more carefully.

\section{Conclusion}
\label{sec:conclusion}

We present a framework to learn from a labeled dataset of normal and anomalous trajectories a {\it Signal Temporal Logic} (STL) 
specification
that better discriminates among them. 
In particular, we design a Robust Genetic algorithm (ROGE) that 
combines an {\it Evolutionary Algorithm} (EA) for learning
the structure of the  formula  and the {\it Gaussian Process Upper Confidence Bound} 
algorithm (GP-UCB) for synthesizing the formula parameters. 
We compare ROGE with our previous work \cite{BufoBSBLB14} 
and with the Decision Tree approach presented in~\cite{bombara_decision_2016} on
an anomalous trajectory detection problem of a maritime surveillance system and on an Ineffective Inspiratory Effort example.

A significant difference concerning our previous approach~\cite{BBS14, BufoBSBLB14} is that we avoid  reconstructing a generative statistical model from the dataset. Furthermore, we modified both the structure and parameter optimization procedure of the genetic algorithm, which now relays on the robustness semantics of STL. This structural improvement was necessary considering that a naive application of our previous approach~\cite{BufoBSBLB14} 
directly on datasets performs very poorly. 

We compare our method also with the Decision Tree (DTL4STL) approach 
of~\cite{bombara_decision_2016} obtaining a low misclassification rate and producing smaller and more interpretable STL specifications. Furthermore,  we do not restrict the class of the temporal formula to only \emph{eventually} and \emph{globally} and we allow arbitrary temporal nesting.

Our genetic algorithm can get wrong results if the formulae of the initial generation are chosen entirely randomly. We avoid this behavior by considering simpler formulae from the beginning as a result of the \textsc{generateInitialFormulae} routine. This approach is a form of regularization and resembles the choice of the set of primitive of DTL4STL.

As future work, we plan to develop an iterative method which uses the proposed genetic algorithm and the robustness of STL to reduce the misclassification rate of the generated formula. 
The idea is to use the robustness value of a learned formula $\phi$ to identify the region of the trajectory space $\mathcal{D}$ where the generated formula $\phi$ has a high accuracy, i.e. trajectories whose robustness is greater than a positive threshold $h^+$ or smaller than a negative threshold $h^-$. These thresholds can be identified by reducing the number of false positives or false negatives. We can then train a new STL classifier $\phi'$ on the remaining trajectories, having small robustness for $\phi$. This will create a hierarchical classifier, that first tests on $\phi$,  if robustness is too low it tests on $\phi'$, and so on. The depth of such classification is limited only by the remaining data at each step. The method can be further strengthen by relying on bagging to generate an ensemble of classifiers at each level, averaging their predictions or choosing the best answer, i.e. the one with larger robustness.

\section*{Acknowledgment}

E.B.\ and L.N.\ acknowledge the partial support of the Austrian National 
Research Network  S 11405-N23 (RiSE/SHiNE) of the Austrian Science 
Fund (FWF). E.B.,\ L.N.\ and S.S.\ acknowledge the partial support of the ICT COST Action IC1402 (ARVI).

\bibliographystyle{splncs}
\bibliography{biblio} 

\begin{thebibliography}{10}
\providecommand{\url}[1]{\texttt{#1}}
\providecommand{\urlprefix}{URL }

\bibitem{DTL4STL}
{DTL4STL}. \url{http://sites.bu.edu/hyness/dtl4stl/} (2016)

\bibitem{AckermannCHRSL10}
Ackermann, C., Cleaveland, R., Huang, S., Ray, A., Shelton, C.P., Latronico,
  E.: Automatic requirement extraction from test cases. In: Proc. of {RV} 2010.
  pp. 1--15 (2010)

\bibitem{Angluin87}
Angluin, D.: Learning regular sets from queries and counterexamples. Inf.
  Comput.  75(2),  87--106 (1987)

\bibitem{AnnpureddyLFS11}
Annpureddy, Y., Liu, C., Fainekos, G.E., Sankaranarayanan, S.: S-taliro: {A}
  tool for temporal logic falsification for hybrid systems. In: Proc. of
  {TACAS} 2011. pp. 254--257. Springer (2011)

\bibitem{AsarinDMN11}
Asarin, E., Donz{\'{e}}, A., Maler, O., Nickovic, D.: Parametric identification
  of temporal properties. In: Proc. of {RV}. pp. 147--160 (2012)

\bibitem{BartocciBNS15}
Bartocci, E., Bortolussi, L., Nenzi, L., Sanguinetti, G.: System design of
  stochastic models using robustness of temporal properties. Theor. Comput.
  Sci.  587,  3--25 (2015)

\bibitem{BBS14}
Bartocci, E., Bortolussi, L., Sanguinetti, G.: Data-driven statistical learning
  of temporal logic properties. In: Proc. of {FORMATS}. pp. 23--37 (2014)

\bibitem{bombara_decision_2016}
Bombara, G., Vasile, C.I., Penedo, F., Yasuoka, H., Belta, C.: A {Decision}
  {Tree} {Approach} to {Data} {Classification} {Using} {Signal} {Temporal}
  {Logic}. In: Proc. of {HSCC}. pp. 1--10 (2016)

\bibitem{BufoBSBLB14}
Bufo, S., Bartocci, E., Sanguinetti, G., Borelli, M., Lucangelo, U.,
  Bortolussi, L.: Temporal logic based monitoring of assisted ventilation in
  intensive care patients. In: Proc. of ISoLA. pp. 391--403 (2014)

\bibitem{Chen2013}
Chen, Y., Tumova, J., Ulusoy, A., Belta, C.: Temporal logic robot control based
  on automata learning of environmental dynamics. The International Journal of
  Robotics Research  32(5),  547--565 (2013)

\bibitem{Donze2013}
Donz{\'e}, A., Ferrer, T., Maler, O.: Efficient robust monitoring for stl. In:
  Proc. of {CAV}. pp. 264--279 (2013)

\bibitem{Donze10}
Donz{\'{e}}, A.: Breach, {A} toolbox for verification and parameter synthesis
  of hybrid systems. In: Proc. of {CAV} 2010. pp. 167--170. Springer (2010)

\bibitem{FuTHC14}
Fu, J., Tanner, H.G., Heinz, J., Chandlee, J.: Adaptive symbolic control for
  finite-state transition systems with grammatical inference. {IEEE} Trans.
  Automat. Contr.  59(2),  505--511 (2014)

\bibitem{HoxhaDF18}
Hoxha, B., Dokhanchi, A., Fainekos, G.E.: Mining parametric temporal logic
  properties in model-based design for cyber-physical systems. {STTT}  20(1),
  79--93 (2018)

\bibitem{IsbernerHS14}
Isberner, M., Howar, F., Steffen, B.: The {TTT} algorithm: {A} redundancy-free
  approach to active automata learning. In: Proc. of {RV} 2014. pp. 307--322
  (2014)

\bibitem{JinDDS15}
Jin, X., Donz{\'{e}}, A., Deshmukh, J.V., Seshia, S.A.: Mining requirements
  from closed-loop control models. {IEEE} Trans. on {CAD} of Integrated
  Circuits and Systems  34(11),  1704--1717 (2015)

\bibitem{kong_temporal_2017}
Kong, Z., Jones, A., Belta, C.: Temporal {Logics} for {Learning} and
  {Detection} of {Anomalous} {Behavior}. IEEE Transactions on Automatic Control
   62(3),  1210--1222 (Mar 2017)

\bibitem{Maler2004}
Maler, O., Nickovic, D.: Monitoring temporal properties of continuous signals.
  In: Proc. of FORMATS. LNCS, vol. 3253, pp. 152--166 (2004)

\bibitem{MalerN13}
Maler, O., Nickovic, D.: Monitoring properties of analog and mixed-signal
  circuits. {STTT}  15(3),  247--268 (2013)

\bibitem{nguyen_abnormal_2017}
Nguyen, L.V., Kapinski, J., Jin, X., Deshmukh, J.V., Butts, K., Johnson, T.T.:
  Abnormal {Data} {Classification} {Using} {Time}-{Frequency} {Temporal}
  {Logic}. In: Proc. of {HSCC}. pp. 237--242 (2017)

\bibitem{Rasmussen2007}
Rasmussen, C.E., Williams, C.K.I.: Gaussian Processes for Machine Learning. MIT
  Press (2006)

\bibitem{Sankaranarayanan17}
Sankaranarayanan, S., Kumar, S.A., Cameron, F., Bequette, B.W., Fainekos, G.E.,
  Maahs, D.M.: Model-based falsification of an artificial pancreas control
  system. {SIGBED} Review  14(2),  24--33 (2017)

\bibitem{SilvettiPB17}
Silvetti, S., Policriti, A., Bortolussi, L.: An active learning approach to the
  falsification of black box cyber-physical systems. In: Proc. of {IFM}. pp.
  3--17 (2017)

\bibitem{gpucb}
Srinivas, N., Krause, A., Kakade, S.M., Seeger, M.W.: Information-theoretic
  regret bounds for gaussian process optimization in the bandit setting. IEEE
  Transactions on Information Theory  58(5),  3250--3265 (2012)

\bibitem{SteffenHI12}
Steffen, B., Howar, F., Isberner, M.: Active automata learning: From dfas to
  interface programs and beyond. In: Proc. of {ICGI} 2012. pp. 195--209 (2012)

\bibitem{Xu2018}
Xu, Z., Julius, A.A.: Census signal temporal logic inference for multiagent
  group behavior analysis. IEEE Transactions on Automation Science and
  Engineering  15(1),  264--277 (2018)

\bibitem{ZhouRWT17}
Zhou, J., Ramanathan, R., Wong, W., Thiagarajan, P.S.: Automated property
  synthesis of odes based bio-pathways models. In: Proc. of {CMSB}. pp.
  265--282 (2017)

\bibitem{0001SDKJ15}
Zutshi, A., Sankaranarayanan, S., Deshmukh, J.V., Kapinski, J., Jin, X.:
  Falsification of safety properties for closed loop control systems. In: Proc.
  of HSCC. pp. 299--300 (2015)

\end{thebibliography}


\end{document}